\documentclass[sigconf]{acmart}

\usepackage{booktabs} 
\usepackage{hyperref}
\usepackage{url}
\usepackage{derivative}
\usepackage{soul}
\usepackage{graphicx}
\usepackage{amsmath}
\usepackage{mathtools}
\usepackage{xspace}
\usepackage{booktabs}
\usepackage{xcolor}
\usepackage{tabularx}
\usepackage{color, colortbl}
\usepackage{diagbox}
\usepackage{multirow}
\usepackage{lipsum}
\usepackage{tabularray}
\usepackage{colortbl}
\usepackage{booktabs}
\usepackage{enumitem}
\usepackage{ragged2e}

\usepackage{graphicx,pifont}
\usepackage[caption=false]{subfig}
\usepackage{parcolumns}
\usepackage{listings}
\usepackage{subfiles}
\usepackage{array}
\usepackage{tabularx,ragged2e}
\usepackage[dvipsnames]{xcolor}

\definecolor{armygreen}{rgb}{0.29, 0.33, 0.13}
\definecolor{darkgreen}{rgb}{0.0, 0.5, 0.0}

\newcommand{\xmark}{\textcolor{OrangeRed}{\ding{55}}}
\newcommand{\tmark}{\textcolor{ForestGreen}{\ding{51}}}

\copyrightyear{2025}
\acmYear{2025}
\setcopyright{cc}
\setcctype{by}
\acmConference[UbiComp Companion '25]{Companion of the 2025 ACM International Joint Conference on Pervasive and Ubiquitous Computing}{October 12--16, 2025}{Espoo, Finland}
\acmBooktitle{Companion of the 2025 ACM International Joint Conference on Pervasive and Ubiquitous Computing (UbiComp Companion '25), October 12--16, 2025, Espoo, Finland}\acmDOI{10.1145/3714394.3756187}
\acmISBN{979-8-4007-1477-1/2025/10}

\begin{document}

\title{LLaSA: A Sensor-Aware LLM for Natural Language Reasoning of Human Activity from IMU Data}

\author{Sheikh Asif Imran Shouborno}
\orcid{0000-0002-2403-3080}
\affiliation{%
\institution{Worcester Polytechnic Institute}
\department{Electrical and Computer Engineering}
\city{Worcester}
\state{Massachusetts}
\country{United States}}
\email{simran@wpi.edu}

\author{Mohammad Nur Hossain Khan}
\orcid{0000-0002-8312-7001}
\affiliation{%
\institution{Worcester Polytechnic Institute}
\department{Electrical and Computer Engineering}
\city{Worcester}
\state{Massachusetts}
\country{United States}}
\email{mkhan@wpi.edu}

\author{Subrata Biswas}
\orcid{0000-0002-2670-0115}
\affiliation{%
\institution{Worcester Polytechnic Institute}
\department{Electrical and Computer Engineering}
\city{Worcester}
\state{Massachusetts}
\country{United States}}
\email{sbiswas@wpi.edu}

\author{Bashima Islam}
\orcid{0000-0002-1917-054X}
\affiliation{%
\institution{Worcester Polytechnic Institute}
\department{Electrical and Computer Engineering}
\city{Worcester}
\state{Massachusetts}
\country{United States}}
\email{bislam@wpi.edu}

\begin{abstract}
Wearable systems can recognize activities from IMU data but often fail to explain their underlying causes or contextual significance. To address this limitation, we introduce two large-scale resources: \textbf{SensorCap}, comprising 35,960 IMU--caption pairs, and \textbf{OpenSQA}, with 199,701 question--answer pairs designed for causal and explanatory reasoning. OpenSQA includes a curated tuning split (Tune-OpenSQA) optimized for scientific accuracy, narrative clarity, and diagnostic insight. Leveraging these datasets, we develop \textbf{LLaSA} (Large Language and Sensor Assistant), a family of compact sensor-aware language models (7B and 13B) that generate interpretable, context-rich responses to open-ended questions grounded in raw IMU data. LLaSA outperforms commercial LLMs, including GPT-3.5 and GPT-4o-mini, on benchmark and real-world tasks, demonstrating the effectiveness of domain supervision and model alignment for sensor reasoning.
\end{abstract}



\maketitle 

\section{Introduction}

\begin{figure}
    \centering
    \includegraphics[width=0.5\textwidth]{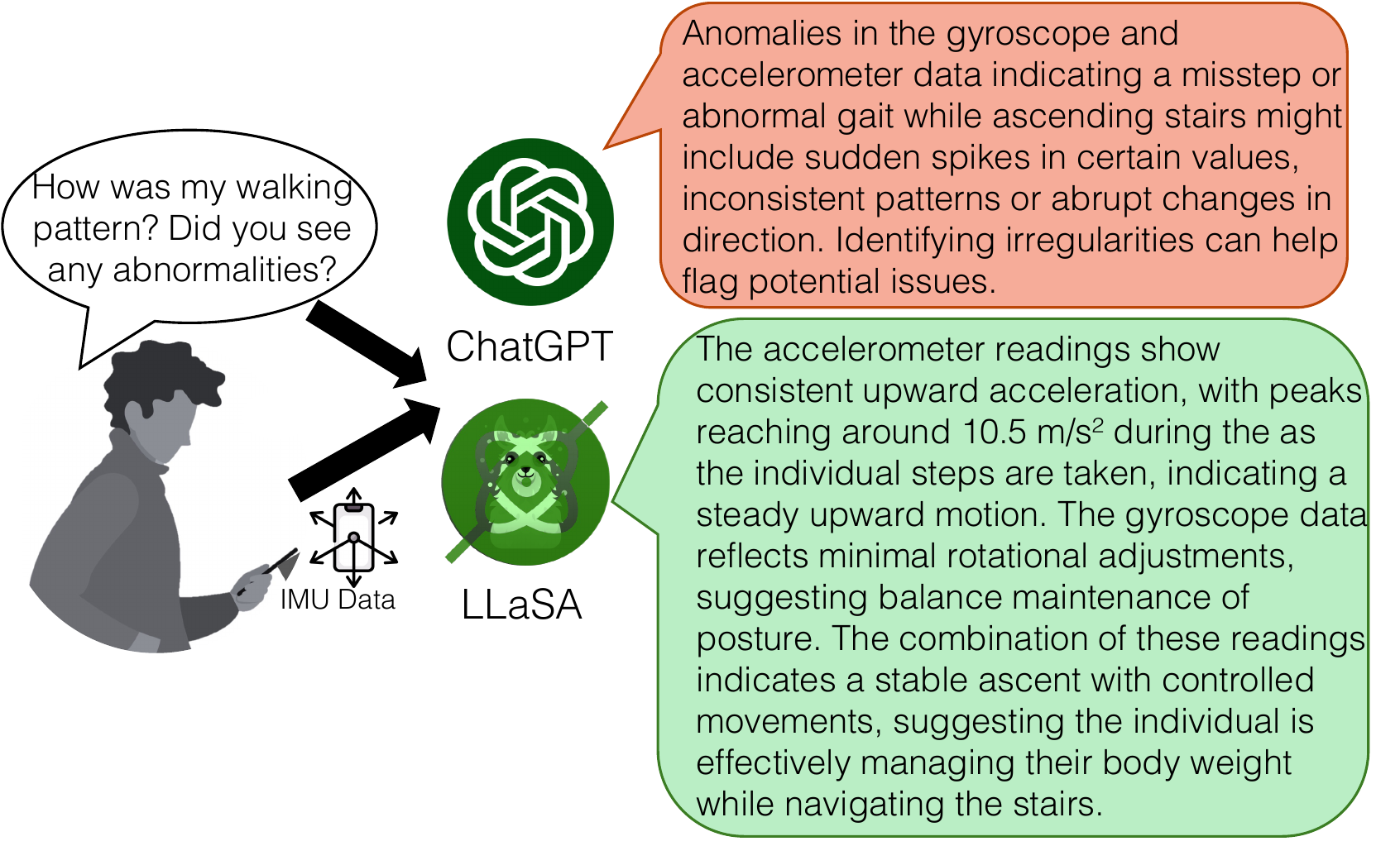}
    \vspace{-2em}
    \caption{A \textbf{real example} comparing GPT-3.5 and LLaSA on an open-ended walking-pattern query. GPT-3.5 responds generically, while LLaSA grounds its explanation in IMU data—referencing upward acceleration and gyroscope stability to infer balanced movement.}
    \label{fig:intro}
\end{figure}

Imagine asking: \textit{“Did I show signs of abnormal gait while climbing stairs? If so, why?”} A meaningful response grounded in motion data could provide real-time insights without technical expertise. Despite over 19 billion IoT devices~\cite{19billion}, 73\% of wearable IMU data remains underused due to its complexity~\cite{newelectronics}. Enabling access to this underutilized sensor data can enhance decision-making across health, fitness, and rehabilitation domains.

While large language models (LLMs) have advanced multimodal understanding in vision and video~\cite{wang2023chatvideo, bai2023qwen}, sensor-based question answering (SQA) remains underexplored. IMU signals are dense, noisy, and lack inherent semantic structure, making alignment with language challenging. Prior models~\cite{chen2021deep, li2024sensorllm} focus on activity classification and label prediction, limiting their ability to explain behavior or reason over time.

We address this gap by introducing two large-scale datasets for open-ended reasoning over IMU data. \textbf{SensorCap} provides 35,960 IMU–caption pairs describing motion, while \textbf{OpenSQA} includes 199,701 question–answer pairs focused on causal and explanatory reasoning. A curated tuning fold of 19,440 examples promotes scientific accuracy, coherence, and diagnostic relevance.

Using these resources, we train LLaSA‑7B and LLaSA‑13B, multimodal LLMs for open-ended, context-aware  QA grounded in IMU data. LLaSA employs a sensor-aware encoder and projection module to compress high-frequency signals into a shared representation that fits within a 2048-token limit.

Figure~\ref{fig:intro} shows LLaSA’s advantage: Unlike GPT-3.5, LLaSA grounds its response in specific IMU signals, referencing upward acceleration and gyroscopic stability to explain a controlled stair ascent, demonstrating causal, sensor-grounded reasoning.

\noindent\textbf{Our key contributions are:}
\begin{itemize}
\item \textbf{Datasets:} We release \textit{SensorCap} and \textit{OpenSQA}, establishing a benchmark for sensor–language reasoning. The tuning fold supports optimization for explanation quality.
\item \textbf{Method:} We propose a GPT-based tuning strategy that promotes scientific accuracy, narrative coherence, and response reliability, without relying solely on classification labels.
\item \textbf{Model:} We introduce \textit{LLaSA}, the first open-source LLM designed for open-ended reasoning over raw IMU data, capable of zero-shot classification and explanatory QA in real-world scenarios.
\end{itemize}

LLaSA is evaluated against GPT‑3.5, Vicuna‑13B, and GPT‑4o‑mini on scientific depth, narrative quality, and contextual grounding. It outperforms GPT‑3.5 and Vicuna‑13B and surpasses GPT‑4o‑mini in sensor-aware QA. Human evaluations (n=22) show strong preference for LLaSA. Real-world trials with 10 users demonstrate generalization to unseen activities and zero-shot classification. Ablations reveal that lower LoRA ranks improve alignment, and GPT-based scoring outperforms loss-based tuning for quality estimation.
\section{Sensor-Language Datasets}\label{sec:dataset}

Understanding motion from IMU data requires more than activity classification. It involves contextual reasoning, narrative interpretation, and causal explanation. Existing datasets are largely limited to discrete labels, offering little support for natural language interaction or open-ended tasks. We introduce two large-scale datasets, \textbf{SensorCap} and \textbf{OpenSQA}, that enable training and evaluation of language models for sensor-based reasoning. All datasets are publicly available.\footnote{\href{https://github.com/BASHLab/LLaSA}{https://github.com/BASHLab/LLaSA}}

\subsection{Source HAR Datasets}

Both SensorCap and OpenSQA are derived from six publicly available human activity recognition (HAR) datasets and one dataset we collected in the wild, as summarized in Table~\ref{tab:opensqa_sources}. The in-the-wild dataset introduces natural noise and variability, allowing us to assess model robustness, reliability, and real-world applicability in the test set. We recruited 10 adults through campus bulletin boards and social media. Each participant used their personal smartphone to record sensor data with the Sensor Logger app~\cite{tszheichoiSensorLogger} while performing \textit{nine} predefined activities, each lasting two minutes (activity list shown in Table~\ref{tab:opensqa_sources}). Recordings took place in diverse environments—approximately 70\% on campus and 30\% at home. Devices varied across iPhone, Samsung, and Google Pixel models, capturing a wide range of hardware and sensor characteristics.

\subsection{SensorCap: Sensor Data Captioning}

SensorCap consists of 35,960 IMU–caption pairs. Each sample includes raw accelerometer and gyroscope signals, sensor location, derived features (e.g., FFT, min, max, median, and variance) and the activity label. The numerical data are downsampled to 10 Hz and truncated to 6 digits after the decimal point.  For each instance, we generate two types of natural language outputs: (1) \textit{a summary caption of at most 25 words describing dominant motion patterns}, and (2) \textit{a step-by-step narration of up to 500 tokens explaining sensor dynamics over time}.

To generate SensorCap, we prompted GPT-4o-mini with characteristic IMU features (e.g., FFT, maxima, variance), raw gyroscope and accelerometer data, and a brief activity summary. To input sensor data into GPT-4o-mini, we converted time-series IMU readings (accelerometer and gyroscope) into structured JSON-like arrays with up to 100 time steps. Additional features such as FFT, statistical summaries (mean, max, min, variance), and sensor locations were appended as human-readable feature descriptors. This structured input was paired with an activity label and passed as the context to GPT-4o-mini. The model generated multi-paragraph explanations grounded in both temporal signal patterns and derived statistical features. For example, the input:

\texttt{Summary: biking; Gyroscope: [[...]]; Accelerometer: [[...]]; Features: FFT, max, min, median, moving avg, variance}

produces a multi-paragraph response narrating rotational and linear acceleration patterns consistent with biking. The model identified z-axis gravitational force, lateral motion for turns, and gyroscopic peaks suggesting leaning behavior and terrain adaptation. The narration also referenced oscillations, balance corrections, and exertion cues tied to real-world biking dynamics.

These structured narrations form the input for our prompt suite, which supports OpenSQA question-answer generation. Each prompt conditions GPT-4o-mini with SensorCap-derived descriptions and guides it to produce QA pairs grounded in sensor reasoning across dimensions such as biomechanics, reliability, and contextual interpretation.

\begin{table}[t]
\caption{Datasets are used to generate SensorCap and OpenSQA. Bolded activities are unseen during training \& tuning. Total 188,429 activities from 95 users.}
\label{tab:opensqa_sources}
\begin{footnotesize}
\begin{tabular}{|l|p{4cm}|c|c|}
\hline
\textbf{Dataset} & \textbf{Classes} & \textbf{Train} & \textbf{Test} \\
\hline
HHAR \cite{stisen2015smart}& bike, sit, stand, walk, ascend stairs, descend stairs & \tmark & \xmark  \\
\hline
Motion \cite{malekzadeh2019mobile} & sit, stand, ascend stairs, descend~stairs, walk, jog & \tmark & \xmark \\
\hline
Shoaib \cite{shoaib2014fusion}& sit, stand, jog, bike, ascend stairs, descend~stairs, walk & \tmark & \xmark  \\
\hline
UCI \cite{reyes2016transition}& stand, sit, lay, walk, ascend stairs, descend stairs & \tmark & \xmark \\
\hline
PAMAP2 \cite{chowdhery2023palm} & lay, sit, stand, walk, run, cycle, Nordic walk, watch TV, computer work, etc. & \tmark & \tmark  \\
\hline
In-the-Wild& sit, stand, \textbf{pick object}, jump, \textbf{swirl}, \textbf{stretch}, use stair, \textbf{elevator}, walk & \xmark  & \tmark \\
\hline
\end{tabular}
\end{footnotesize}
\end{table}

\subsection{OpenSQA: Sensor-Aware QA Dataset}

OpenSQA extends SensorCap to support open-ended question answering grounded in IMU data. It contains 199,701 question–answer pairs, each generated using gyroscope and accelerometer signals, activity labels, sensor locations, and SensorCap-provided motion features and narrations. Each QA pair is a natural language response to a user-style question grounded in this multimodal sensor context. For each instance, we prompt GPT-4o-mini to generate five QA pairs spanning factual (e.g., \textit{“What activity is being performed?”}), descriptive (e.g., \textit{“What trends are visible?”}), causal (e.g., \textit{“Why is this motion occurring?”}), diagnostic (e.g., \textit{“Is this motion stable or noisy?”}), and inferential (e.g., \textit{“What might happen next?”}) reasoning. A subset of generations is refined with human feedback to improve clarity and contextual alignment. 

To support structured evaluation beyond label prediction, we partition OpenSQA into two folds: A \textbf{training fold} includes \textbf{179,727} QA pairs constructed from five datasets: HHAR, MotionSense, Shoaib, UCI HAR, and the training portion of PAMAP2~\cite{chowdhery2023palm}, and a \textbf{tuning fold}, \textbf{Tune-OpenSQA}, contains 19,440 QA pairs generated from a held-out PAMAP2 subset with GPT-4o. It serves as a benchmark for evaluating sensor-aware reasoning across scientific consistency, narrative clarity, and signal reliability.

For further real-world assessment, we include an in-the-wild dataset of \textbf{534 QA pairs}. This subset evaluates model generalization under diverse environmental conditions and device variation.

\subsection{Data Quality Assessment}

We assess dataset quality using three LLMs as automatic judges: Deepseek-V3 (accessed via API), LLaMA-3.1-8B, and Gemma-12B (run locally). Each model scores question–answer pairs on a 1-5 scale based on alignment with IMU data, activity label, and narration. Deepseek-V3 rates 87\% of samples above 4.5 and 96\% above 4.0. Gemma-12B scores 73\% above 4.5 and 93\% above 4.0. LLaMA-3.1-8B gives 97\% above 4.5 and 100\% above 4.0. The consistency across models confirms the high quality of the generated data and its suitability for sensor-grounded reasoning tasks.
\section{Large Language and Sensor Assistant}
Large Language and Sensor Assistant (LLaSA), available in 7B and 13B variants, processes raw IMU data and user queries to generate grounded responses for versatile wearable sensing applications.

\begin{figure}[t]
\centering
\includegraphics[width=0.5\textwidth]{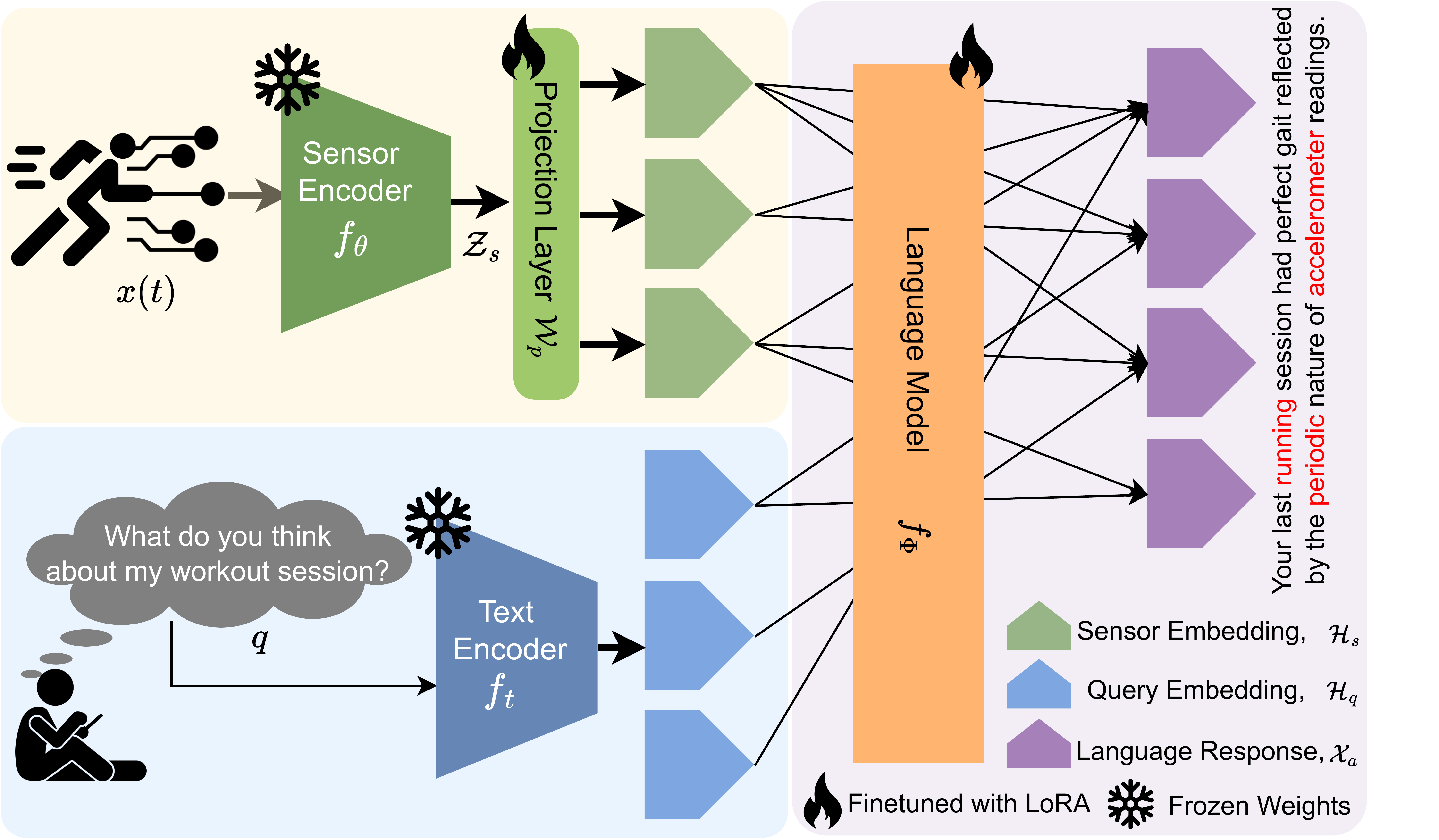}\
\caption{\textbf{LLaSA Overview:} Raw IMU data is encoded by $f_{\theta}$, projected into a shared space $\mathcal{H}_s$, and fused with query embeddings $\mathcal{H}_q$ before being processed by $f_{\Phi}$ to generate a response $\mathcal{X}_a$.}
\label{fig:architecture}
\end{figure}

\subsection{LLaSA Architecture}
Figure~\ref{fig:architecture} illustrates the LLaSA architecture, which includes a sensor encoder for feature extraction, a query encoder, a projection module for multimodal alignment, and a language model conditioned on both inputs.

\noindent\textbf{Sensor Encoder ${f_{\theta}}$.}
Given IMU signals $x(t) \in \mathbb{R}^{C \times T}$, the frozen encoder $f_{\theta}$ produces: $\mathcal{Z}_s = f_{\theta}(x(t)) \in \mathbb{R}^{\bar{c} \times \bar{k}}$. 
We adopt LIMU-BERT~\cite{xu2021limu}, which uses self-attention to encode temporal and channel-wise patterns in $\mathcal{Z}_s$.

\noindent\textbf{Query Encoder $f_t(\cdot)$.}
User queries $Q$ are tokenized using SentencePiece~\cite{kudo1808sentencepiece} (LLaMA-compatible), producing embeddings $\mathcal{H}_q \in \mathbb{R}^{\bar{c} \times k}$.

\noindent\textbf{Projection and Alignment Layer.}
Following previous works \cite{liu2023llava}, you leverage $2$ layers of fully connected layers, along side \textit{GELU}-activation to project the encoded sensor vector $\mathcal{Z}_s$ to sensor embedding $\mathcal{H}_s = \sigma(\mathcal{W}_p \mathcal{Z}_s)$, where $ \mathcal{W}_p \in \mathbb{R}^{\bar{k} \times k}$ is the learnable parameters of projection and alignment layer.

Then we concatenate sensor and text embeddings to form the fused input $\mathcal{H}_{\text{input}} = [\mathcal{H}_s; \mathcal{H}_q]$
This projection acts as a sensor-aware tokenizer, enabling cross-modal consistency.

\noindent\textbf{Language Model $f_{\Phi}$.} The LLM generates output $\mathcal{X}_a$ through autoregressive decoding by modeling the conditional probability as $P(\mathcal{X}_a | \mathcal{H}_{\text{input}}) = \prod_{t=1}^{T} P(x_{a,t} | \mathcal{H}_{\text{input}}, x_{a,<t}; \phi)$ where $\phi$ represents the model parameters. Conditioning on the sensor history $\mathcal{H}_s$ allows for grounded, context-aware reasoning over IMU inputs.

\subsection{LLaSA Hyperparameter Tuning}
\label{sec:hyperparam}

Integrating sensor data into LLMs presents challenges beyond standard text-based tuning. Sensor-aware models must process noisy, continuous signals, preserve temporal dependencies, and generate responses grounded in physical context. Prior work often relies on classification labels, limiting a model’s ability to generate explanatory outputs.
We propose a tuning strategy that improves interpretability and reliability without modifying model architecture. It combines targeted hyperparameter search with a structured evaluation protocol to support causal reasoning and multimodal alignment.

\noindent\textbf{Hyperparameter Search Strategy.}
We tune five key parameters to balance convergence and expressivity: learning rate ($\eta$), batch size ($B$), weight decay ($\lambda$), LoRA rank ($r$), and dropout rate ($p_{\text{drop}}$). Following prior work~\cite{sellam2020bleurt}, we select configurations using the evaluation metric below to ensure fidelity in sensor-grounded responses.

\begin{table}[]
\caption{BLEU/ROUGE reward generic overlap. GPT scoring reflects reasoning and relevance in sensor-grounded QA.}
\label{table:evaluation_comparison}
\resizebox{\linewidth}{!}{
\begin{tabular}{@{}|l|l|l|@{}}
\toprule
Query &
  Generic Response &
  Informed Response \\ \midrule
\begin{tabular}[c]{@{}l@{}}What is the\\ person doing?\end{tabular} &
  \begin{tabular}[c]{@{}l@{}}"The person is moving."\\ \textbf{BLEU/ROUGE:} High \\ \textbf{GPT:} Low\end{tabular} &
  \begin{tabular}[c]{@{}l@{}}"Smooth gait suggests walking."\\ \textbf{BLEU/ROUGE:} Medium \\ \textbf{GPT:} High\end{tabular} \\ \midrule
\begin{tabular}[c]{@{}l@{}}Is the activity\\ classification\\ correct?\end{tabular} &
  \begin{tabular}[c]{@{}l@{}}"Yes, it matches the label."\\ \textbf{BLEU/ROUGE:} High \\ \textbf{GPT:} Low\end{tabular} &
  \begin{tabular}[c]{@{}l@{}}"Patterns match walking signature."\\ \textbf{BLEU/ROUGE:} Medium \\ \textbf{GPT:} High\end{tabular} \\ \bottomrule
\end{tabular}
}
\end{table}

\noindent\textbf{Structured LLM-Based Evaluation Metric.}
Lexical metrics like BLEU~\cite{papineni2002bleu} and ROUGE~\cite{lin-2004-rouge} often fail to reflect semantic or causal relevance, especially in sensor-grounded QA where phrasing varies~\cite{gao2024llm, desmond2024evalullm}. As shown in Table~\ref{table:evaluation_comparison}, vague answers like \textit{“The person is moving”} can receive high BLEU scores despite lacking depth.

To address this, we score answers using GPT-4o along four axes: \textit{correctness}, alignment with sensor dynamics; \textit{completeness}, coverage of key observations; \textit{consistency}, logical flow; and \textit{helpfulness}, clarity and informativeness. Each is rated on a 0–100 scale, and the average yields a final score:
\begin{equation*}
Q_{\text{final}} = \frac{Q_{\text{correctness}} + Q_{\text{completeness}} + Q_{\text{consistency}} + Q_{\text{helpfulness}}}{4}
\end{equation*}

This scoring captures coherence, context, and reasoning better than lexical overlap. For instance, \textit{“Smooth gait suggests walking”} scores higher than generic responses. We validate the approach with Deepseek-V3, observing strong inter-rater agreement (ICC = 0.92, \textbf{p} = 0.0001), confirming its reliability.

\subsection{Implementation Details}

\noindent\textbf{LIMU-BERT Training.} \label{limubert-training}
We pretrain LIMU-BERT on unlabeled sensor data from HHAR~\cite{stisen2015smart}, UCI-HAR~\cite{reyes2016transition}, MotionSense~\cite{malekzadeh2019mobile}, and Shoaib~\cite{shoaib2014fusion}, covering 8 activities with 3-axis accelerometer and gyroscope signals downsampled to 20 Hz. The best checkpoint is selected after ~8 hours of training (up to 1000 epochs) with Adam and early stopping on an RTX 3090 Ti. Additional PAMAP2 activities~\cite{roggen2010collecting} yielded negligible gains, indicating sufficient diversity in the core datasets.

\noindent\textbf{Multimodal Projector Training.}
We train a GELU-activated MLP ($W_P$) on OpenSQA to align LIMU-BERT sensor features with token embeddings. Training uses 16-bit precision, Adam optimizer, learning rate $1\text{e}^{-4}$, and batch size 32 on A100 GPUs. One epoch suffices for convergence.

\noindent\textbf{LLaSA Fine-Tuning.}
We fine-tune Vicuna-7B and 13B with LoRA~\cite{hu2021lora}, using the pretrained projector and LLaVA-style~\cite{liu2023llava} instruction tuning. Sensor embeddings fit within a 2048-token context window. The 7B model is trained on one A100 GPU; 13B uses two.

Fine-tuning runs for one epoch (batch size 16) with LoRA ranks $r \in \{4, 8, 16, 64, 128\}$ and scaling $\alpha = r$~\cite{lee2023platypus}. We sweep learning rates of $1\text{e}^{-4}$, $3\text{e}^{-5}$, $1\text{e}^{-6}$ for LoRA, and $2\text{e}^{-5}$, $1\text{e}^{-4}$, $1\text{e}^{-3}$ for the projector.

\section{Evaluation for Open-Ended Reasoning Question Answering}

We evaluate LLaSA’s ability to generate interpretable, context-rich answers from raw IMU data. Unlike standard classification, this task involves open-ended questions that require causal reasoning and contextual understanding. Our evaluation framework includes three components: (1) a structured LLM-based evaluation metric (Section~\ref{sec:hyperparam}), (2) human satisfaction ratings, and (3) generalization to in-the-wild settings with unseen activities that test LLaSA’s ability to reason about novel motion patterns. We also assess LLaSA’s performance on zero-shot human activity classification. While LLaSA uses normalized 20 Hz samples to generate sensor tokens, others receive them as texts downsampled to 10 Hz and truncated up to 6 digits after the decimal point to minimize the number of tokens.

\begin{table}[]
\centering
\caption{LLaSA leads in science score and overall quality, outperforming stronger-context LLMs.}
\label{table:quality_scores}
\resizebox{\linewidth}{!}{
\begin{tabular}{|l|c|c|c|c|c|}
\hline
\textbf{Model} & \textbf{Context Len.} & \textbf{Science} & \textbf{Narration} & \textbf{Reliability} & \textbf{Overall} \\ \hline
LLaSA & 2k & \textbf{75.0} & 71.0 & 70.5 & 72.17 \\ \hline
GPT-3.5-Turbo & 16k & 64.5 & 65.0 & 60.0 & 63.17 \\ \hline
GPT-4o-mini & 128k & 72.0 & \textbf{77.0} & \textbf{78.0} & 75.67 \\ \hline
Vicuna-13b & 16k & 63.4 & 67.0 & 65.4 & 65.27 \\ \hline
\end{tabular}
}]
\end{table}

\subsection{Comparison with General LLMs}

We compare LLaSA‑13B to GPT‑3.5, GPT‑4o‑mini, and Vicuna‑13B using a structured evaluation of scientific reasoning, narrative clarity, and reliability (Table~\ref{table:quality_scores}). All models are scored on 30 QA pairs from the OpenSQA tuning fold (PAMAP2 data).

LLaSA outperforms GPT‑3.5 and Vicuna‑13B across all metrics, producing more sensor-aware and causally grounded answers. It leads in scientific reasoning, highlighting the value of domain-specific tuning. GPT‑4o‑mini performs well but lacks depth in physics reasoning, showing general models benefit from domain adaptation.

Despite a 2K context limit, LLaSA matches or exceeds models with up to 64$\times$ larger context windows. Its strong performance—without architectural changes—demonstrates that careful tuning can enable compact models to reason accurately over sensor data.

\begin{figure}[]
\centering
\includegraphics[width=0.5\textwidth]{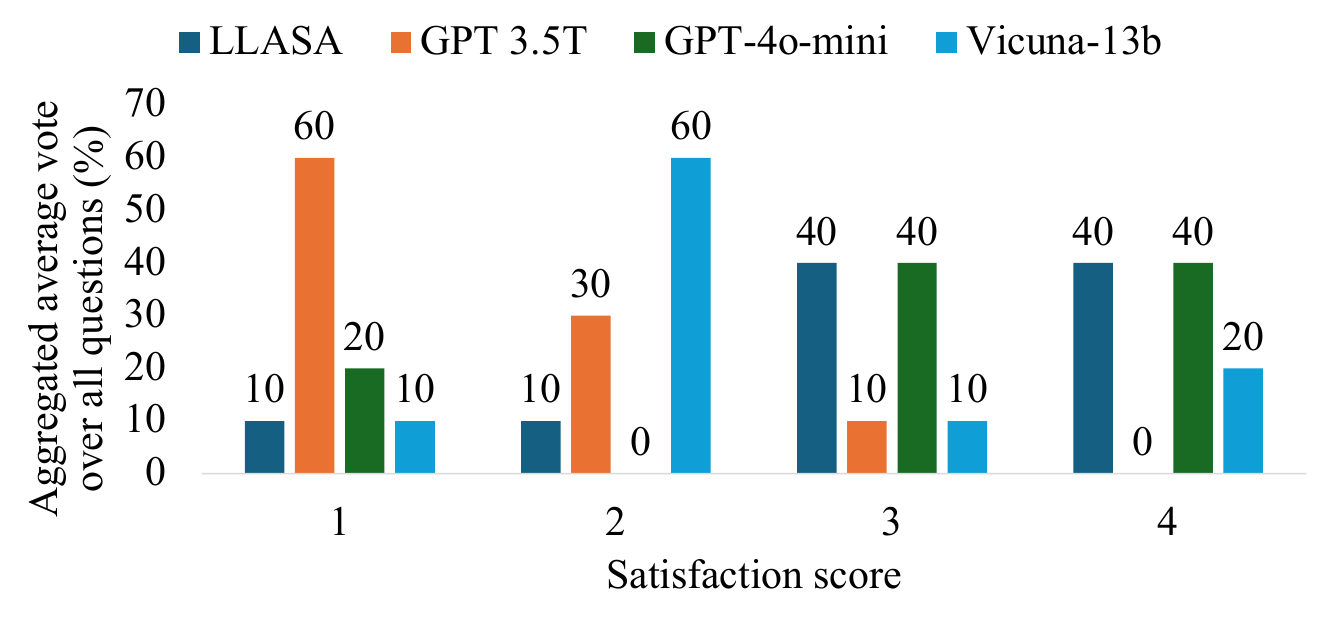}
\caption{LLaSA yields more consistent high human satisfaction scores than baselines.}
\label{fig:survey_result}
\end{figure}

\begin{figure*}[!htb]
\centering
\includegraphics[width=\textwidth]{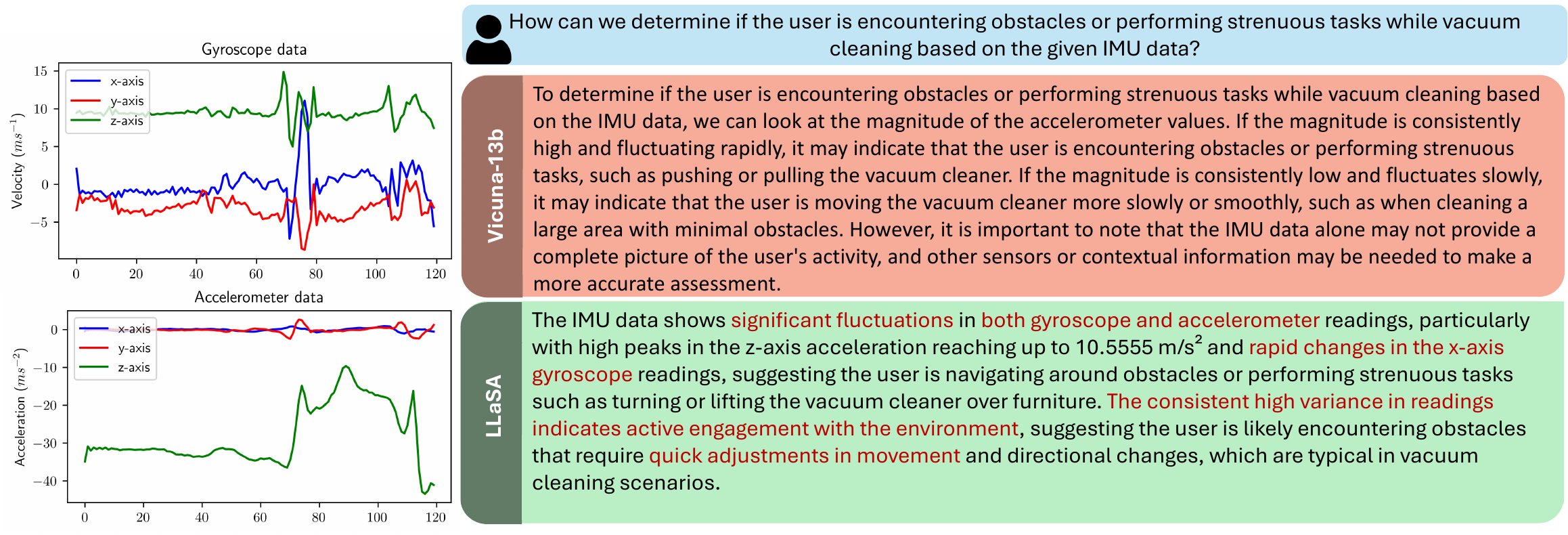}
\caption{Example showing LLaSA provides enriched, contextual answers (\textbf{highlighted in red}) based on IMU input.}
\label{fig:qa_example}
\end{figure*}

\subsection{Human Evaluation of QA Quality}

We conducted a user study with 21 participants to evaluate the perceived quality of sensor-based answers. Each participant reviewed 10 QA examples covering scientific reasoning, narrative clarity, and reliability. Questions were paired with anonymized outputs from LLaSA-13B, GPT-3.5, GPT-4o-mini, and Vicuna-13B. Visual plots and brief IMU descriptions aided interpretation.

Participants ranked answers from 1 (worst) to 4 (best) based on clarity, relevance, and grounding. LLaSA scored highest in 40\% of cases, matching GPT-4o-mini, but with fewer low scores, showing greater consistency. Figure~\ref{fig:survey_result} highlights LLaSA’s precision in interpreting subtle sensor cues.

Figure~\ref{fig:qa_example} shows an example: while Vicuna-13B offers a generic explanation, LLaSA cites specific signal fluctuations—e.g., z-axis acceleration and gyroscope variance—to infer obstacles and strenuous activity during vacuum cleaning.

Participants noted that LLaSA improved their understanding by linking low-level signals to high-level explanations. Even non-experts found its answers more informative and accessible, reinforcing the value of domain-specific tuning for real-world use.

\begin{table}[]
\centering
\caption{LLaSA-7B delivers more coherent and reliable responses than Vicuna-13B on in-the-wild data.}
\resizebox{\linewidth}{!}{
\begin{tabular}{|l|c|c|c|c|c|}
\hline
\textbf{Model} & 
\textbf{Model Size} & 
\textbf{Context Length} & 
\textbf{Narrative} & \textbf{Reliability} & \textbf{Overall} \\ \hline
Vicuna & 13B & 16K & 72.0 & 78.0 & 75.0 \\ \hline
LLaSA & 7B & 2K & \textbf{78.2} & \textbf{81.7} & \textbf{79.95}\\
\hline
\end{tabular}%
}
\label{tab:realword_results}
\end{table}

\subsection{Generalization to In-the-Wild Data}
We evaluate LLaSA-7B on an OpenSQA test fold that includes only in-the-wild samples collected using smartphones in real-world environments. This evaluation assesses narrative quality and reliability under noise and device variations.

We compare LLaSA-7B with Vicuna-13B-1.5-16K, a significantly larger general-purpose model. To ensure fairness, GPT-4o scores the outputs using prompts designed to account for variability in real-world sensor signals. As shown in Table~\ref{tab:realword_results}, LLaSA delivers more coherent and reliable answers across a wide range of activities, despite its smaller model size and limited context window. These results highlight LLaSA’s ability to generalize beyond curated datasets and maintain performance in practical deployment scenarios.

\begin{table*}[]
\caption{Zero-shot classification performance on human activity recognition. LLaSA outperforms GPT models and Vicuna-7B with 2.6–12× higher F1-scores, despite using 8–64$\times$ smaller context windows.}
\vspace{-1em}
\label{tab:closeendedresults}
\resizebox{0.99\linewidth}{!}{
\begin{tabular}{|@{}l|c|ccc|ccc|ccc|ccc|ccc@{}|}
\toprule
\multicolumn{1}{|c|}{\multirow{2}{*}{LMA}}              &\multicolumn{1}{c}{\multirow{2}{*}{\begin{tabular}{@{}l@{}}
                   Context \\
                   length
              \end{tabular}}}  &\multicolumn{3}{|c|}{HHAR}                   & \multicolumn{3}{c|}{MotionSense} & \multicolumn{3}{c|}{Shoaib} & \multicolumn{3}{c|}{UCI}   & \multicolumn{3}{c|}{SHL}   \\ \cmidrule(l){3-17}
\multicolumn{1}{|c|}{}                &\multicolumn{1}{c|}{}  & F-1  & Precision & Recall & F-1    & Precision   & Recall   & F-1   & Precision & Recall & F-1  & Precision & Recall & F-1  & Precision & Recall \\ \hline
Vicuna-7b-16K & 16k            & 0.05 & 0.12      & 0.03   & 0.09   & 0.15        & 0.12     & 0.08  & 0.11      & 0.11   & 0.04 & 0.25      & 0.03   & 0.08 & 0.38      & 0.07   \\ \hline
GPT-3.5-T     & 16k            & 0.07 & 0.16      & 0.1    & 0.08   & 0.07        & 0.13     & 0.09  & 0.1       & 0.12   & 0.07 & 0.35      & 0.12   & 0.15 & 0.22      & 0.15   \\ \hline
GPT-4o-mini   & 128k           & 0.16 & 0.24      & 0.18   & 0.12   & 0.25        & 0.15     & 0.1   & 0.19      & 0.13   & 0.15 & 0.17      & 0.13   & 0.25 & 0.36      & 0.23   \\ \hline
GPT-4o        & 128k           & 0.07 & 0.14      & 0.08   & 0.08   & 0.26        & 0.06     & 0.14  & 0.35      & 0.15   & 0.04 & 0.16      & 0.03   & 0.22 & 0.27      & 0.23   \\ \hline
GPT-3.5-T-F   & 16k            & 0.52 & 0.54      & 0.7    & 0.21   & 0.27        & 0.21     & 0.27  & 0.45      & 0.31   & 0.24 & 0.28      & 0.31   & 0.23 & 0.25      & 0.32   \\ \hline
LLaSA         & \textbf{2k}             & \textbf{0.84} & \textbf{0.88}      & \textbf{0.86}   & \textbf{0.83}   & \textbf{0.85}        & \textbf{0.83}     & \textbf{0.81}  & \textbf{0.84}      & \textbf{0.81}   & \textbf{0.72} & \textbf{0.82}      & \textbf{0.75}   & \textbf{0.65} & \textbf{0.76}      & \textbf{0.71} \\ \bottomrule 
\end{tabular}}

\end{table*}

\subsection{Zero-Shot Activity Classification}

We evaluate LLaSA’s ability to perform zero-shot activity recognition, where the model classifies IMU signals without task-specific fine-tuning by selecting the most plausible activity label.

We use five datasets: HHAR, Motion, Shoaib, UCI, and SHL~\cite{gjoreski2018university}. Each sample includes derived IMU features and candidate activity labels (Table~\ref{tab:opensqa_sources}). The model is prompted to choose the correct label; if none is returned, the output is marked “Unclear.” Performance is measured using precision, recall, and F1-score.

Compared to Vicuna-7B, GPT-3.5, GPT-4o-mini, and a fine-tuned GPT-3.5 (GPT-3.5-T-F), LLaSA achieves 2.6–12$\times$ higher F1 across all datasets (Table~\ref{tab:closeendedresults}). It generalizes well even to SHL, unseen during training. Despite a 2K token limit, LLaSA outperforms models with up to 128K context length, demonstrating strong efficiency and domain grounding.

\begin{table}[]
\centering
\caption{Hyperparameter tuning and model scaling help find the best-performing model through scientific, narrative, and reliability assessment QA Tasks.}
\vspace{-1em}
\label{table:hyperparams}
\resizebox{\linewidth}{!}{
\begin{tabular}{|c|c|c|c|c||c|c|c|c|}
\hline
\begin{tabular}[c]{@{}c@{}}Model\\ size\end{tabular} & \begin{tabular}[c]{@{}c@{}}\#QA\end{tabular} & LR                    & \begin{tabular}[c]{@{}c@{}}Projector\\ LR\end{tabular} & \begin{tabular}[c]{@{}c@{}}LORA\\ Rank\end{tabular} & \begin{tabular}[c]{@{}c@{}}Science \\ QA Quality\end{tabular} & \begin{tabular}[c]{@{}c@{}}Narration\\ QA Quality\end{tabular} & \begin{tabular}[c]{@{}c@{}}Reliability\\ QA Quality\end{tabular} & \begin{tabular}[c]{@{}c@{}}Overall\\ Score\end{tabular} \\ \hline
\multirow{22}{*}{7b}                                 & \multirow{12}{*}{131,407}                                 & \multirow{6}{*}{1e-4} & \multirow{3}{*}{1e-4}                                  & 8    & 41.0                                                          & 52.5                                                           & 45.0                                                             & 46.1                                                    \\
                                                     &                                                           &                       &                                                        & 16   & 32.5                                                          & 54.0                                                           & 44.5                                                             & 43.7                                                    \\
                                                     &                                                           &                       &                                                        & 128  & 35.5                                                          & 49.0                                                           & 44.0                                                             & 42.8                                                    \\ \cline{4-9} 
                                                     &                                                           &                       & \multirow{3}{*}{2e-5}                                  & 8    & 32.0                                                          & \textbf{57.0}                                                  & 42.0                                                             & 43.7                                                    \\
                                                     &                                                           &                       &                                                        & 16   & 36.0                                                          & 55.0                                                           & 42.0                                                             & 44.3                                                    \\
                                                     &                                                           &                       &                                                        & 128  & 35.0                                                          & 34.0                                                           & 48.0                                                             & 39.0                                                    \\ \cline{3-9} 
                                                     &                                                           & \multirow{6}{*}{3e-5} & \multirow{3}{*}{1e-4}                                  & 8    & \textbf{56.0}                                                 & 56.5                                                           & 51.5                                                             & \textbf{54.7}                                           \\
                                                     &                                                           &                       &                                                        & 16   & 49.5                                                          & 53.5                                            
               & 52.0                                                             & 51.7                                                    \\
                                                     &                                                           &                       &                                                        & 128  & 44.0                                                          & 49.0                                                           & 44.5                                                             & 45.8                                                    \\ \cline{4-9} 
                                                     &                                                           &                       & \multirow{3}{*}{2e-5}                                  & 8    & 50.0                                                          & 54.0                                                           & 52.5                                                             & 52.2                                                    \\
                                                     &                                                           &                       &                                                        & 16   & 29.0                                                          & 42.0                                                           & 49.5                                                             & 40.2                                                    \\
                                                     &                                                           &                       &                                                        & 128  & 42.0                                                          & 50.0                                                           & 49.5                                                             & 47.2                                                    \\ \cline{2-9} 
                                                     & \multirow{10}{*}{179,727}                                 & \multirow{3}{*}{1e-4} & \multirow{3}{*}{1e-4}                                  & 8    & 34.5                                                          & 43.0                                                           & 46.0                                                             & 41.2                                                    \\
                                                     &                                                           &                       &                                                        & 16   & 27.5                                                          & 40.5                                                           & \textbf{58.5}                                                    & 42.2                                                    \\
                                                     &                                                           &                       &                                                        & 128  & 32.0                                                          & 38.0                                                           & 53.5                                                             & 41.2                                                    \\ \cline{3-9} 
                                                     &                                                           & \multirow{4}{*}{1e-6} & \multirow{2}{*}{1e-3}                                  & 4    & 36.5                                                          & 45.5                                                           & 34.5                                                             & 38.8                                                    \\
                                                     &                                                           &                       &                                                        & 8    & 32.5                                                          & 39.5                                                           & 32.5                                                             & 34.8                                                    \\ \cline{4-9} 
                                                     &                                                           &                       & \multirow{2}{*}{1e-4}                                  & 4    & 35.0                                                          & 45.0                                                           & 47.0                                                             & 42.3                                                    \\
                                                     &                                                           &                       &                                                        & 8    & 30.0                                                          & 43.0                                                           & 32.5                                                             & 35.2                                                    \\ \cline{3-9} 
                                                     &                                                           & \multirow{3}{*}{3e-5} & \multirow{2}{*}{1e-3}                                  & 4    & 28.0                                                          & 40.5                                                           & 50.0                                                             & 39.5                                                    \\
                                                     &                                                           &                       &                                                        & 8    & 31.0                                                          & 38.5                                            
               & 48.5                                                             & 38.2                                                    \\ \cline{4-9} 
                                                     &                                                           &                       & 1e-4                                                   & 4    & 34.0                                                          & 44.0                                                           & 49.0                                                             & 42.3                                                    \\ \hline
13b                                                  & 179,727                                                   & 3e-5                  & 1e-4                                                   & 8    & \textbf{75.0}                                                 & \textbf{71.0}                                                  & \textbf{70.5}                                                    & \textbf{72.17}                                          \\ \hline
\end{tabular}
}
\end{table}

\section{Tuning Strategy Evaluation and Insights}
We study how LoRA rank, model size, and dataset scale influence performance. LLaSA is fine-tuned on 30 QA examples spanning diverse reasoning types, varying LoRA rank, learning rate, and model size (Table~\ref{table:hyperparams}). Rank 8 consistently performs best across science, narration, and reliability, supporting prior findings that low-rank adaptation improves alignment and reduces overfitting on noisy IMU data. The 13B model outperforms 7B across all metrics, though both benefit from tuning. Expanding to the full PAMAP2 QA dataset (179k examples) offers marginal improvement, suggesting saturation beyond a certain scale.

Notably, the lowest-loss model (1.474) performs worse than the best one (1.526), confirming that training loss is an unreliable proxy for QA quality. Our structured quality scores better reflect interpretability, grounding, and reasoning, and are more suitable for tuning sensor-aware language models.

\section{Related Work}

Recent research has explored leveraging large language models for sensor-grounded reasoning tasks. Penetrative AI~\cite{xu2024penetrative} demonstrated that models like ChatGPT can interpret textualized IoT sensor data to perform perception and activity inference tasks in physical domains. RAVEN~\cite{biswas2025raven} introduces a cross-modal gating mechanism over audio, video, and sensor streams, showing significant improvement in multimodal QA accuracy by weighting relevant modalities.

In the sensor-Language alignment space, LLMSense~\cite{ouyang2024llmsense} incorporates long-term spatiotemporal sensor traces through selective history inclusion and summarization before reasoning on the cloud, achieving over 80\% accuracy in tasks like dementia diagnosis and occupancy tracking. SensorLLM~\cite{zhang2025sensorlm} aligns sensor channel data with LLMs using special tokens and trend-descriptive text, enabling robust activity recognition without manual annotations.

Other systems—such as IMUGPT2.0~\cite{leng2024imugpt}, Sensor2Text~\cite{chen2024sensor2text}, and MentalLLM~\cite{xu2024mental}—focus on generating summaries or labels from sensor data, but lack the open‑ended reasoning and QA functionality that LLaSA supports.

A unique approach called By My Eyes~\cite{yoon-etal-2024-eyes} converts IMU and ECG signals to visual prompts for multimodal LLMs, whereas LLaSA directly models raw sequences without modality conversion

In healthcare, models like Health‑LLM~\cite{kim2024health}, MedTSLLM~\cite{chan2024medtsllm}, PH‑LLM~\cite{cosentino2024towards}, and PHIA~\cite{merrill2024transforming} fine-tune LLMs on wearable sensor data for structured clinical prediction. Surveys such as Zhang et al.~\cite{zhang2024large} review LLM applications to time series, but highlight that most existing methods focus on task-level prediction or forecasting rather than temporally grounded explanations or QA.

\section{Limitations}
One limitation of our approach is the reliance on a single LLM, GPT-4o-mini, for both SensorCap generation and OpenSQA QA synthesis. While GPT-4o offers high-quality reasoning, its performance and tokenization biases inherently influence the quality of the datasets. While inter-rater reliability between GPT and Deepseek was good (0.92), evaluation using GPT-based automatic metrics also inherits this dependency, which may affect reproducibility as models evolve over time.
Another limitation is the scale of human evaluation and benchmark QA comparison. The current evaluation spans 30 QA pairs for benchmarking and 10 examples per human rater, which may limit statistical robustness. Expanding this in future work will help validate the generalization of LLaSA.
We also observe that LLaSA occasionally fails in ambiguous or under-informative input scenarios, such as stationary readings misclassified as “standing” vs. “sitting.” It may also hallucinate motion patterns when the signal lacks clear transitions, or overfit to statistically common movement descriptors even when context differs. Addressing such failure modes is an avenue for future improvement, especially with fine-grained signal conditioning or counterfactual prompts.
Finally, the current model sizes (7B and 13B with 16-bit precision) are unsuitable for real-time inference on mobile or edge devices. Supporting deployment on smartphones or wearables would require model compression strategies such as quantization, distillation, or architecture pruning. Exploring efficient sensor-language reasoning under these constraints is an important direction for future work, particularly for privacy-preserving, on-device applications.

\section{Conclusion}
We present SensorCaps and OpenSQA, two new resources that enable sensor-grounded question answering with open-ended, causal reasoning over IMU data. Built on these datasets, LLaSA shows how lightweight adaptation and structured evaluation can produce interpretable, context-rich answers from raw sensor signals. LLaSA outperforms general-purpose LLMs up to 64$\times$ larger while remaining efficient and real-time capable. These results highlight the value of domain-specific supervision for aligning language models with low-level sensor data. Future work will extend this foundation to support multimodal fusion, interactive dialogue, and longer contexts for deployment in health, HCI, and wearable systems.

\section*{Acknowledgment}
This research was supported by funding from the NSF CNS-2347692. We gratefully acknowledge their support in enabling this work.

\bibliographystyle{ACM-Reference-Format}
\balance
\bibliography{ref}


\begin{thebibliography}{38}


\ifx \showCODEN    \undefined \def \showCODEN     #1{\unskip}     \fi
\ifx \showISBNx    \undefined \def \showISBNx     #1{\unskip}     \fi
\ifx \showISBNxiii \undefined \def \showISBNxiii  #1{\unskip}     \fi
\ifx \showISSN     \undefined \def \showISSN      #1{\unskip}     \fi
\ifx \showLCCN     \undefined \def \showLCCN      #1{\unskip}     \fi
\ifx \shownote     \undefined \def \shownote      #1{#1}          \fi
\ifx \showarticletitle \undefined \def \showarticletitle #1{#1}   \fi
\ifx \showURL      \undefined \def \showURL       {\relax}        \fi
\providecommand\bibfield[2]{#2}
\providecommand\bibinfo[2]{#2}
\providecommand\natexlab[1]{#1}
\providecommand\showeprint[2][]{arXiv:#2}

\bibitem[new(2022)]%
        {newelectronics}
 \bibinfo{year}{2022}\natexlab{}.
\newblock \bibinfo{title}{The {IoT} data deluge in industry and manufacturing}.
\newblock \bibinfo{howpublished}{\url{https://www.newelectronics.co.uk/content/features/the-iot-data-deluge-in-industry-and-manufacturing}}.
\newblock
\newblock
\shownote{Accessed: 2025-2-1}.


\bibitem[Atil et~al\mbox{.}(2024)]%
        {atil2024non}
\bibfield{author}{\bibinfo{person}{Berk Atil}, \bibinfo{person}{Sarp Aykent}, \bibinfo{person}{Alexa Chittams}, \bibinfo{person}{Lisheng Fu}, \bibinfo{person}{Rebecca~J Passonneau}, \bibinfo{person}{Evan Radcliffe}, \bibinfo{person}{Guru~Rajan Rajagopal}, \bibinfo{person}{Adam Sloan}, \bibinfo{person}{Tomasz Tudrej}, \bibinfo{person}{Ferhan Ture}, {et~al\mbox{.}}} \bibinfo{year}{2024}\natexlab{}.
\newblock \showarticletitle{Non-determinism of" deterministic" llm settings}.
\newblock \bibinfo{journal}{\emph{arXiv preprint arXiv:2408.04667}} (\bibinfo{year}{2024}).
\newblock


\bibitem[Bai et~al\mbox{.}(2023)]%
        {bai2023qwen}
\bibfield{author}{\bibinfo{person}{Jinze Bai}, \bibinfo{person}{Shuai Bai}, \bibinfo{person}{Shusheng Yang}, \bibinfo{person}{Shijie Wang}, \bibinfo{person}{Sinan Tan}, \bibinfo{person}{Peng Wang}, \bibinfo{person}{Junyang Lin}, \bibinfo{person}{Chang Zhou}, {and} \bibinfo{person}{Jingren Zhou}.} \bibinfo{year}{2023}\natexlab{}.
\newblock \showarticletitle{Qwen-vl: A frontier large vision-language model with versatile abilities}.
\newblock \bibinfo{journal}{\emph{arXiv preprint arXiv:2308.12966}} (\bibinfo{year}{2023}).
\newblock


\bibitem[Biswas et~al\mbox{.}(2025)]%
        {biswas2025raven}
\bibfield{author}{\bibinfo{person}{Subrata Biswas}, \bibinfo{person}{Mohammad Nur~Hossain Khan}, {and} \bibinfo{person}{Bashima Islam}.} \bibinfo{year}{2025}\natexlab{}.
\newblock \showarticletitle{RAVEN: Representational Alignment of Audio, Video, Embedded Sensors and Natural Language}.
\newblock \bibinfo{journal}{\emph{arXiv preprint}} (\bibinfo{year}{2025}).
\newblock


\bibitem[Chan et~al\mbox{.}(2024)]%
        {chan2024medtsllm}
\bibfield{author}{\bibinfo{person}{Nimeesha Chan}, \bibinfo{person}{Felix Parker}, \bibinfo{person}{William Bennett}, \bibinfo{person}{Tianyi Wu}, \bibinfo{person}{Mung~Yao Jia}, \bibinfo{person}{James Fackler}, {and} \bibinfo{person}{Kimia Ghobadi}.} \bibinfo{year}{2024}\natexlab{}.
\newblock \showarticletitle{Medtsllm: Leveraging llms for multimodal medical time series analysis}.
\newblock \bibinfo{journal}{\emph{arXiv preprint arXiv:2408.07773}} (\bibinfo{year}{2024}).
\newblock


\bibitem[Chen et~al\mbox{.}(2021)]%
        {chen2021deep}
\bibfield{author}{\bibinfo{person}{Kaixuan Chen}, \bibinfo{person}{Dalin Zhang}, \bibinfo{person}{Lina Yao}, \bibinfo{person}{Bin Guo}, \bibinfo{person}{Zhiwen Yu}, {and} \bibinfo{person}{Yunhao Liu}.} \bibinfo{year}{2021}\natexlab{}.
\newblock \showarticletitle{Deep learning for sensor-based human activity recognition: Overview, challenges, and opportunities}.
\newblock \bibinfo{journal}{\emph{ACM Computing Surveys (CSUR)}} \bibinfo{volume}{54}, \bibinfo{number}{4} (\bibinfo{year}{2021}), \bibinfo{pages}{1--40}.
\newblock


\bibitem[Chen et~al\mbox{.}(2024)]%
        {chen2024sensor2text}
\bibfield{author}{\bibinfo{person}{Wenqiang Chen}, \bibinfo{person}{Jiaxuan Cheng}, \bibinfo{person}{Leyao Wang}, \bibinfo{person}{Wei Zhao}, {and} \bibinfo{person}{Wojciech Matusik}.} \bibinfo{year}{2024}\natexlab{}.
\newblock \showarticletitle{Sensor2Text: Enabling Natural Language Interactions for Daily Activity Tracking Using Wearable Sensors}.
\newblock \bibinfo{journal}{\emph{Proceedings of the ACM on Interactive, Mobile, Wearable and Ubiquitous Technologies}} \bibinfo{volume}{8}, \bibinfo{number}{4} (\bibinfo{year}{2024}), \bibinfo{pages}{1--26}.
\newblock


\bibitem[Choi(2022)]%
        {tszheichoiSensorLogger}
\bibfield{author}{\bibinfo{person}{Kelvin Choi}.} \bibinfo{year}{2022}\natexlab{}.
\newblock \bibinfo{title}{Sensor Logger}.
\newblock \bibinfo{howpublished}{\url{https://www.tszheichoi.com/sensorlogger}}.
\newblock
\newblock
\shownote{[Accessed 12-11-2024]}.


\bibitem[Chowdhery et~al\mbox{.}(2023)]%
        {chowdhery2023palm}
\bibfield{author}{\bibinfo{person}{Aakanksha Chowdhery}, \bibinfo{person}{Sharan Narang}, \bibinfo{person}{Jacob Devlin}, \bibinfo{person}{Maarten Bosma}, \bibinfo{person}{Gaurav Mishra}, \bibinfo{person}{Adam Roberts}, \bibinfo{person}{Paul Barham}, \bibinfo{person}{Hyung~Won Chung}, \bibinfo{person}{Charles Sutton}, \bibinfo{person}{Sebastian Gehrmann}, {et~al\mbox{.}}} \bibinfo{year}{2023}\natexlab{}.
\newblock \showarticletitle{Palm: Scaling language modeling with pathways}.
\newblock \bibinfo{journal}{\emph{Journal of Machine Learning Research}} \bibinfo{volume}{24}, \bibinfo{number}{240} (\bibinfo{year}{2023}), \bibinfo{pages}{1--113}.
\newblock


\bibitem[Cosentino et~al\mbox{.}(2024)]%
        {cosentino2024towards}
\bibfield{author}{\bibinfo{person}{Justin Cosentino}, \bibinfo{person}{Anastasiya Belyaeva}, \bibinfo{person}{Xin Liu}, \bibinfo{person}{Nicholas~A Furlotte}, \bibinfo{person}{Zhun Yang}, \bibinfo{person}{Chace Lee}, \bibinfo{person}{Erik Schenck}, \bibinfo{person}{Yojan Patel}, \bibinfo{person}{Jian Cui}, \bibinfo{person}{Logan~Douglas Schneider}, {et~al\mbox{.}}} \bibinfo{year}{2024}\natexlab{}.
\newblock \showarticletitle{Towards a Personal Health Large Language Model}.
\newblock \bibinfo{journal}{\emph{arXiv preprint arXiv:2406.06474}} (\bibinfo{year}{2024}).
\newblock


\bibitem[Desmond et~al\mbox{.}(2024)]%
        {desmond2024evalullm}
\bibfield{author}{\bibinfo{person}{Michael Desmond}, \bibinfo{person}{Zahra Ashktorab}, \bibinfo{person}{Qian Pan}, \bibinfo{person}{Casey Dugan}, {and} \bibinfo{person}{James~M Johnson}.} \bibinfo{year}{2024}\natexlab{}.
\newblock \showarticletitle{EvaluLLM: LLM assisted evaluation of generative outputs}. In \bibinfo{booktitle}{\emph{Companion Proceedings of the 29th International Conference on Intelligent User Interfaces}}. \bibinfo{pages}{30--32}.
\newblock


\bibitem[Gao et~al\mbox{.}(2024)]%
        {gao2024llm}
\bibfield{author}{\bibinfo{person}{Mingqi Gao}, \bibinfo{person}{Xinyu Hu}, \bibinfo{person}{Jie Ruan}, \bibinfo{person}{Xiao Pu}, {and} \bibinfo{person}{Xiaojun Wan}.} \bibinfo{year}{2024}\natexlab{}.
\newblock \showarticletitle{Llm-based nlg evaluation: Current status and challenges}.
\newblock \bibinfo{journal}{\emph{arXiv preprint arXiv:2402.01383}} (\bibinfo{year}{2024}).
\newblock


\bibitem[Gjoreski et~al\mbox{.}(2018)]%
        {gjoreski2018university}
\bibfield{author}{\bibinfo{person}{Hristijan Gjoreski}, \bibinfo{person}{Mathias Ciliberto}, \bibinfo{person}{Lin Wang}, \bibinfo{person}{Francisco Javier~Ordonez Morales}, \bibinfo{person}{Sami Mekki}, \bibinfo{person}{Stefan Valentin}, {and} \bibinfo{person}{Daniel Roggen}.} \bibinfo{year}{2018}\natexlab{}.
\newblock \showarticletitle{The university of sussex-huawei locomotion and transportation dataset for multimodal analytics with mobile devices}.
\newblock \bibinfo{journal}{\emph{IEEE Access}}  \bibinfo{volume}{6} (\bibinfo{year}{2018}), \bibinfo{pages}{42592--42604}.
\newblock


\bibitem[Hu et~al\mbox{.}(2021)]%
        {hu2021lora}
\bibfield{author}{\bibinfo{person}{Edward~J Hu}, \bibinfo{person}{Yelong Shen}, \bibinfo{person}{Phillip Wallis}, \bibinfo{person}{Zeyuan Allen-Zhu}, \bibinfo{person}{Yuanzhi Li}, \bibinfo{person}{Shean Wang}, \bibinfo{person}{Lu Wang}, {and} \bibinfo{person}{Weizhu Chen}.} \bibinfo{year}{2021}\natexlab{}.
\newblock \showarticletitle{Lora: Low-rank adaptation of large language models}.
\newblock \bibinfo{journal}{\emph{arXiv preprint arXiv:2106.09685}} (\bibinfo{year}{2021}).
\newblock


\bibitem[Kim et~al\mbox{.}(2024)]%
        {kim2024health}
\bibfield{author}{\bibinfo{person}{Yubin Kim}, \bibinfo{person}{Xuhai Xu}, \bibinfo{person}{Daniel McDuff}, \bibinfo{person}{Cynthia Breazeal}, {and} \bibinfo{person}{Hae~Won Park}.} \bibinfo{year}{2024}\natexlab{}.
\newblock \showarticletitle{Health-llm: Large language models for health prediction via wearable sensor data}.
\newblock \bibinfo{journal}{\emph{arXiv preprint arXiv:2401.06866}} (\bibinfo{year}{2024}).
\newblock


\bibitem[Kudo and Richardson(1808)]%
        {kudo1808sentencepiece}
\bibfield{author}{\bibinfo{person}{Taku Kudo} {and} \bibinfo{person}{John Richardson}.} \bibinfo{year}{1808}\natexlab{}.
\newblock \showarticletitle{Sentencepiece: A simple and language independent subword tokenizer and detokenizer for neural text processing. arXiv 2018}.
\newblock \bibinfo{journal}{\emph{arXiv preprint arXiv:1808.06226}} (\bibinfo{year}{1808}).
\newblock


\bibitem[Lee et~al\mbox{.}(2023)]%
        {lee2023platypus}
\bibfield{author}{\bibinfo{person}{Ariel~N Lee}, \bibinfo{person}{Cole~J Hunter}, {and} \bibinfo{person}{Nataniel Ruiz}.} \bibinfo{year}{2023}\natexlab{}.
\newblock \showarticletitle{Platypus: Quick, cheap, and powerful refinement of llms}.
\newblock \bibinfo{journal}{\emph{arXiv preprint arXiv:2308.07317}} (\bibinfo{year}{2023}).
\newblock


\bibitem[Leng et~al\mbox{.}(2024)]%
        {leng2024imugpt}
\bibfield{author}{\bibinfo{person}{Zikang Leng}, \bibinfo{person}{Amitrajit Bhattacharjee}, \bibinfo{person}{Hrudhai Rajasekhar}, \bibinfo{person}{Lizhe Zhang}, \bibinfo{person}{Elizabeth Bruda}, \bibinfo{person}{Hyeokhyen Kwon}, {and} \bibinfo{person}{Thomas Pl{\"o}tz}.} \bibinfo{year}{2024}\natexlab{}.
\newblock \showarticletitle{Imugpt 2.0: Language-based cross modality transfer for sensor-based human activity recognition}.
\newblock \bibinfo{journal}{\emph{Proceedings of the ACM on Interactive, Mobile, Wearable and Ubiquitous Technologies}} \bibinfo{volume}{8}, \bibinfo{number}{3} (\bibinfo{year}{2024}), \bibinfo{pages}{1--32}.
\newblock


\bibitem[Li et~al\mbox{.}(2024)]%
        {li2024sensorllm}
\bibfield{author}{\bibinfo{person}{Zechen Li}, \bibinfo{person}{Shohreh Deldari}, \bibinfo{person}{Linyao Chen}, \bibinfo{person}{Hao Xue}, {and} \bibinfo{person}{Flora~D Salim}.} \bibinfo{year}{2024}\natexlab{}.
\newblock \showarticletitle{Sensorllm: Aligning large language models with motion sensors for human activity recognition}.
\newblock \bibinfo{journal}{\emph{arXiv preprint arXiv:2410.10624}} (\bibinfo{year}{2024}).
\newblock


\bibitem[Lin(2004)]%
        {lin-2004-rouge}
\bibfield{author}{\bibinfo{person}{Chin-Yew Lin}.} \bibinfo{year}{2004}\natexlab{}.
\newblock \showarticletitle{{ROUGE}: A Package for Automatic Evaluation of Summaries}. In \bibinfo{booktitle}{\emph{Text Summarization Branches Out}}. \bibinfo{publisher}{Association for Computational Linguistics}, \bibinfo{address}{Barcelona, Spain}, \bibinfo{pages}{74--81}.
\newblock
\urldef\tempurl%
\url{https://aclanthology.org/W04-1013/}
\showURL{%
\tempurl}


\bibitem[Liu et~al\mbox{.}(2023)]%
        {liu2023llava}
\bibfield{author}{\bibinfo{person}{Shilong Liu}, \bibinfo{person}{Hao Cheng}, \bibinfo{person}{Haotian Liu}, \bibinfo{person}{Hao Zhang}, \bibinfo{person}{Feng Li}, \bibinfo{person}{Tianhe Ren}, \bibinfo{person}{Xueyan Zou}, \bibinfo{person}{Jianwei Yang}, \bibinfo{person}{Hang Su}, \bibinfo{person}{Jun Zhu}, {et~al\mbox{.}}} \bibinfo{year}{2023}\natexlab{}.
\newblock \showarticletitle{Llava-plus: Learning to use tools for creating multimodal agents}.
\newblock \bibinfo{journal}{\emph{arXiv preprint arXiv:2311.05437}} (\bibinfo{year}{2023}).
\newblock


\bibitem[Malekzadeh et~al\mbox{.}(2019)]%
        {malekzadeh2019mobile}
\bibfield{author}{\bibinfo{person}{Mohammad Malekzadeh}, \bibinfo{person}{Richard~G Clegg}, \bibinfo{person}{Andrea Cavallaro}, {and} \bibinfo{person}{Hamed Haddadi}.} \bibinfo{year}{2019}\natexlab{}.
\newblock \showarticletitle{Mobile sensor data anonymization}. In \bibinfo{booktitle}{\emph{Proceedings of the international conference on internet of things design and implementation}}. \bibinfo{pages}{49--58}.
\newblock


\bibitem[Merrill et~al\mbox{.}(2024)]%
        {merrill2024transforming}
\bibfield{author}{\bibinfo{person}{Mike~A Merrill}, \bibinfo{person}{Akshay Paruchuri}, \bibinfo{person}{Naghmeh Rezaei}, \bibinfo{person}{Geza Kovacs}, \bibinfo{person}{Javier Perez}, \bibinfo{person}{Yun Liu}, \bibinfo{person}{Erik Schenck}, \bibinfo{person}{Nova Hammerquist}, \bibinfo{person}{Jake Sunshine}, \bibinfo{person}{Shyam Tailor}, {et~al\mbox{.}}} \bibinfo{year}{2024}\natexlab{}.
\newblock \showarticletitle{Transforming wearable data into health insights using large language model agents}.
\newblock \bibinfo{journal}{\emph{arXiv preprint arXiv:2406.06464}} (\bibinfo{year}{2024}).
\newblock


\bibitem[Ouyang and Srivastava(2024)]%
        {ouyang2024llmsense}
\bibfield{author}{\bibinfo{person}{Xiaomin Ouyang} {and} \bibinfo{person}{Mani Srivastava}.} \bibinfo{year}{2024}\natexlab{}.
\newblock \showarticletitle{LLMSense: Harnessing LLMs for High-level Reasoning Over Spatiotemporal Sensor Traces}.
\newblock \bibinfo{journal}{\emph{arXiv preprint arXiv:2403.19857}} (\bibinfo{year}{2024}).
\newblock


\bibitem[Papineni et~al\mbox{.}(2002)]%
        {papineni2002bleu}
\bibfield{author}{\bibinfo{person}{Kishore Papineni}, \bibinfo{person}{Salim Roukos}, \bibinfo{person}{Todd Ward}, {and} \bibinfo{person}{Wei-Jing Zhu}.} \bibinfo{year}{2002}\natexlab{}.
\newblock \showarticletitle{Bleu: a method for automatic evaluation of machine translation}. In \bibinfo{booktitle}{\emph{Proceedings of the 40th annual meeting of the Association for Computational Linguistics}}. \bibinfo{pages}{311--318}.
\newblock


\bibitem[Reyes-Ortiz et~al\mbox{.}(2016)]%
        {reyes2016transition}
\bibfield{author}{\bibinfo{person}{Jorge-L Reyes-Ortiz}, \bibinfo{person}{Luca Oneto}, \bibinfo{person}{Albert Sam{\`a}}, \bibinfo{person}{Xavier Parra}, {and} \bibinfo{person}{Davide Anguita}.} \bibinfo{year}{2016}\natexlab{}.
\newblock \showarticletitle{Transition-aware human activity recognition using smartphones}.
\newblock \bibinfo{journal}{\emph{Neurocomputing}}  \bibinfo{volume}{171} (\bibinfo{year}{2016}), \bibinfo{pages}{754--767}.
\newblock


\bibitem[Roggen et~al\mbox{.}(2010)]%
        {roggen2010collecting}
\bibfield{author}{\bibinfo{person}{Daniel Roggen}, \bibinfo{person}{Alberto Calatroni}, \bibinfo{person}{Mirco Rossi}, \bibinfo{person}{Thomas Holleczek}, \bibinfo{person}{Kilian F{\"o}rster}, \bibinfo{person}{Gerhard Tr{\"o}ster}, \bibinfo{person}{Paul Lukowicz}, \bibinfo{person}{David Bannach}, \bibinfo{person}{Gerald Pirkl}, \bibinfo{person}{Alois Ferscha}, {et~al\mbox{.}}} \bibinfo{year}{2010}\natexlab{}.
\newblock \showarticletitle{Collecting complex activity datasets in highly rich networked sensor environments}. In \bibinfo{booktitle}{\emph{2010 Seventh international conference on networked sensing systems (INSS)}}. IEEE, \bibinfo{pages}{233--240}.
\newblock


\bibitem[Sellam et~al\mbox{.}(2020)]%
        {sellam2020bleurt}
\bibfield{author}{\bibinfo{person}{Thibault Sellam}, \bibinfo{person}{Dipanjan Das}, {and} \bibinfo{person}{Ankur~P Parikh}.} \bibinfo{year}{2020}\natexlab{}.
\newblock \showarticletitle{BLEURT: Learning robust metrics for text generation}.
\newblock \bibinfo{journal}{\emph{arXiv preprint arXiv:2004.04696}} (\bibinfo{year}{2020}).
\newblock


\bibitem[Shoaib et~al\mbox{.}(2014)]%
        {shoaib2014fusion}
\bibfield{author}{\bibinfo{person}{Muhammad Shoaib}, \bibinfo{person}{Stephan Bosch}, \bibinfo{person}{Ozlem~Durmaz Incel}, \bibinfo{person}{Hans Scholten}, {and} \bibinfo{person}{Paul~JM Havinga}.} \bibinfo{year}{2014}\natexlab{}.
\newblock \showarticletitle{Fusion of smartphone motion sensors for physical activity recognition}.
\newblock \bibinfo{journal}{\emph{Sensors}} \bibinfo{volume}{14}, \bibinfo{number}{6} (\bibinfo{year}{2014}), \bibinfo{pages}{10146--10176}.
\newblock


\bibitem[Sinha(2024)]%
        {19billion}
\bibfield{author}{\bibinfo{person}{Satyajit Sinha}.} \bibinfo{year}{2024}\natexlab{}.
\newblock \bibinfo{title}{State of {IoT} 2024: Number of connected {IoT} devices growing 13\% to 18.8 billion globally}.
\newblock \bibinfo{howpublished}{\url{https://iot-analytics.com/number-connected-iot-devices/}}.
\newblock
\newblock
\shownote{Accessed: 2025-2-1}.


\bibitem[Stisen et~al\mbox{.}(2015)]%
        {stisen2015smart}
\bibfield{author}{\bibinfo{person}{Allan Stisen}, \bibinfo{person}{Henrik Blunck}, \bibinfo{person}{Sourav Bhattacharya}, \bibinfo{person}{Thor~Siiger Prentow}, \bibinfo{person}{Mikkel~Baun Kj{\ae}rgaard}, \bibinfo{person}{Anind Dey}, \bibinfo{person}{Tobias Sonne}, {and} \bibinfo{person}{Mads~M{\o}ller Jensen}.} \bibinfo{year}{2015}\natexlab{}.
\newblock \showarticletitle{Smart devices are different: Assessing and mitigatingmobile sensing heterogeneities for activity recognition}. In \bibinfo{booktitle}{\emph{Proceedings of the 13th ACM conference on embedded networked sensor systems}}. \bibinfo{pages}{127--140}.
\newblock


\bibitem[Wang et~al\mbox{.}(2023)]%
        {wang2023chatvideo}
\bibfield{author}{\bibinfo{person}{Junke Wang}, \bibinfo{person}{Dongdong Chen}, \bibinfo{person}{Chong Luo}, \bibinfo{person}{Xiyang Dai}, \bibinfo{person}{Lu Yuan}, \bibinfo{person}{Zuxuan Wu}, {and} \bibinfo{person}{Yu-Gang Jiang}.} \bibinfo{year}{2023}\natexlab{}.
\newblock \showarticletitle{Chatvideo: A tracklet-centric multimodal and versatile video understanding system}.
\newblock \bibinfo{journal}{\emph{arXiv preprint arXiv:2304.14407}} (\bibinfo{year}{2023}).
\newblock


\bibitem[Xu et~al\mbox{.}(2024a)]%
        {xu2024penetrative}
\bibfield{author}{\bibinfo{person}{Huatao Xu}, \bibinfo{person}{Liying Han}, \bibinfo{person}{Qirui Yang}, \bibinfo{person}{Mo Li}, {and} \bibinfo{person}{Mani Srivastava}.} \bibinfo{year}{2024}\natexlab{a}.
\newblock \showarticletitle{Penetrative ai: Making llms comprehend the physical world}. In \bibinfo{booktitle}{\emph{Proceedings of the 25th International Workshop on Mobile Computing Systems and Applications}}. \bibinfo{pages}{1--7}.
\newblock


\bibitem[Xu et~al\mbox{.}(2021)]%
        {xu2021limu}
\bibfield{author}{\bibinfo{person}{Huatao Xu}, \bibinfo{person}{Pengfei Zhou}, \bibinfo{person}{Rui Tan}, \bibinfo{person}{Mo Li}, {and} \bibinfo{person}{Guobin Shen}.} \bibinfo{year}{2021}\natexlab{}.
\newblock \showarticletitle{Limu-bert: Unleashing the potential of unlabeled data for imu sensing applications}. In \bibinfo{booktitle}{\emph{Proceedings of the 19th ACM Conference on Embedded Networked Sensor Systems}}. \bibinfo{pages}{220--233}.
\newblock


\bibitem[Xu et~al\mbox{.}(2024b)]%
        {xu2024mental}
\bibfield{author}{\bibinfo{person}{Xuhai Xu}, \bibinfo{person}{Bingsheng Yao}, \bibinfo{person}{Yuanzhe Dong}, \bibinfo{person}{Saadia Gabriel}, \bibinfo{person}{Hong Yu}, \bibinfo{person}{James Hendler}, \bibinfo{person}{Marzyeh Ghassemi}, \bibinfo{person}{Anind~K Dey}, {and} \bibinfo{person}{Dakuo Wang}.} \bibinfo{year}{2024}\natexlab{b}.
\newblock \showarticletitle{Mental-llm: Leveraging large language models for mental health prediction via online text data}.
\newblock \bibinfo{journal}{\emph{Proceedings of the ACM on Interactive, Mobile, Wearable and Ubiquitous Technologies}} \bibinfo{volume}{8}, \bibinfo{number}{1} (\bibinfo{year}{2024}), \bibinfo{pages}{1--32}.
\newblock


\bibitem[Yoon et~al\mbox{.}(2024)]%
        {yoon-etal-2024-eyes}
\bibfield{author}{\bibinfo{person}{Hyungjun Yoon}, \bibinfo{person}{Biniyam~Aschalew Tolera}, \bibinfo{person}{Taesik Gong}, \bibinfo{person}{Kimin Lee}, {and} \bibinfo{person}{Sung-Ju Lee}.} \bibinfo{year}{2024}\natexlab{}.
\newblock \showarticletitle{By My Eyes: Grounding Multimodal Large Language Models with Sensor Data via Visual Prompting}. In \bibinfo{booktitle}{\emph{Proceedings of the 2024 Conference on Empirical Methods in Natural Language Processing}}, \bibfield{editor}{\bibinfo{person}{Yaser Al-Onaizan}, \bibinfo{person}{Mohit Bansal}, {and} \bibinfo{person}{Yun-Nung Chen}} (Eds.). \bibinfo{publisher}{Association for Computational Linguistics}, \bibinfo{address}{Miami, Florida, USA}, \bibinfo{pages}{2219--2241}.
\newblock
\href{https://doi.org/10.18653/v1/2024.emnlp-main.133}{doi:\nolinkurl{10.18653/v1/2024.emnlp-main.133}}


\bibitem[Zhang et~al\mbox{.}(2024)]%
        {zhang2024large}
\bibfield{author}{\bibinfo{person}{Xiyuan Zhang}, \bibinfo{person}{Ranak~Roy Chowdhury}, \bibinfo{person}{Rajesh~K Gupta}, {and} \bibinfo{person}{Jingbo Shang}.} \bibinfo{year}{2024}\natexlab{}.
\newblock \showarticletitle{Large language models for time series: A survey}.
\newblock \bibinfo{journal}{\emph{arXiv preprint arXiv:2402.01801}} (\bibinfo{year}{2024}).
\newblock


\bibitem[Zhang et~al\mbox{.}(2025)]%
        {zhang2025sensorlm}
\bibfield{author}{\bibinfo{person}{Yuwei Zhang}, \bibinfo{person}{Kumar Ayush}, \bibinfo{person}{Siyuan Qiao}, \bibinfo{person}{A~Ali Heydari}, \bibinfo{person}{Girish Narayanswamy}, \bibinfo{person}{Maxwell~A Xu}, \bibinfo{person}{Ahmed~A Metwally}, \bibinfo{person}{Shawn Xu}, \bibinfo{person}{Jake Garrison}, \bibinfo{person}{Xuhai Xu}, {et~al\mbox{.}}} \bibinfo{year}{2025}\natexlab{}.
\newblock \showarticletitle{SensorLM: Learning the Language of Wearable Sensors}.
\newblock \bibinfo{journal}{\emph{arXiv preprint arXiv:2506.09108}} (\bibinfo{year}{2025}).
\newblock


\end{thebibliography}

\appendix
\setcounter{table}{0}
\renewcommand{\thetable}{A\arabic{table}}
\section*{Appendix A: Prompts and examples}

\begin{table}[H]
\caption{Example from SensorCap: GPT-4o-mini generates a detailed explanation based on summary, sensor data, and statistical features.}
\label{tab:sensorcap_example}
\begin{footnotesize}
\begin{tabular}{|p{0.96\linewidth}|}
\hline
\textbf{Inputs:} \\
Summary: biking \\
Gyroscope: \texttt{[[-0.54474, -0.060323, 0.27657], [-0.437375, -0.952935, 0.26466], [-0.16661, -1.15345, -0.057116], ...]} \\
Accelerometer: \texttt{[[6.97365, -1.498255, -7.5116], [8.0292, -3.24845, -7.8113], [7.647799, -2.5606, -7.27325], ...]} \\
Features: FFT, maximum, minimum, median, moving average, variance \\

\textbf{Outputs:} \\
\textbf{Features:} High rotation rates, significant accelerations, consistent z-axis downward force, noticeable side-to-side motion indicating turns during biking. \\

\textbf{Narration:} The dataset provided showcases IMU sensor readings that detail the nuances of biking. The gyroscope data reveals rotation around the x, y, and z axes, indicating the rider's orientation and turning movements, while the accelerometer readings offer insights into linear accelerations experienced during the biking activity. \\
\textit{Step 1: Analyzing Gyroscope Data.} The values exhibit frequent negative readings along the y-axis, peaking around -1.15, suggesting clockwise turning. Fluctuations in the z-axis indicate tilt variations; x-axis movement is milder. \\
\textit{Step 2: Investigating Accelerometer Data.} A significant downward force averaging around -7.5 reflects steady contact with the ground. Fluctuations suggest moments of acceleration, deceleration, and terrain interaction. \\
\textit{Step 3: Transition Analysis.} Transient gyroscope shifts reflect balance adjustments and terrain response. Peaks in the x-axis correspond to pedaling bursts. \\
\textit{Step 4: Peak and Average Movement.} Oscillatory patterns reflect sharp turns and quick stops. Minimum values align with moments of stabilization before pivoting. \\
\textit{Conclusion.} These signals capture the biker's physical dynamics and balance strategy. The interplay between gyroscopic orientation and linear acceleration builds a detailed profile of biking behavior. \\
\hline
\end{tabular}
\end{footnotesize}
\end{table}

\begin{table}[H]
\caption{Prompt design for generating OpenSQA. GPT-4o-mini is instructed with contextual IMU features to generate reasoning-rich question–answer pairs.}
\label{tab:opensqa_prompt}
\begin{footnotesize}
\resizebox{\linewidth}{!}{
\begin{tabular}{|ll|}
\hline
\multicolumn{2}{|c|}{\cellcolor[HTML]{CBCEFB}\textbf{Instruction to GPT-4o-mini}} \\ \hline
\multicolumn{2}{|p{\columnwidth}|}{You are an expert in interpreting gyroscope and accelerometer data. Based on provided context (including a summary, raw sensor data, statistical features, and a temporal narration), generate five logical question–answer pairs requiring deductive reasoning grounded in the data. Format the output as: \texttt{1. Q: ... A: ...}} \\ \hline
\cellcolor[HTML]{CBCEFB}\textbf{Prompt Input Structure} & \cellcolor[HTML]{CBCEFB}\textbf{Output Structure} \\
\hline
\begin{tabular}[c]{@{}p{0.48\linewidth}@{}}
1. Activity label \\
2. Gyroscope data \\
3. Accelerometer data \\
4. Sensor location \\
5. Extracted IMU features \\
6. Temporal narration (SensorCap)
\end{tabular}
&
Five contextual question–answer pairs \\ \hline
\end{tabular}}
\end{footnotesize}
\end{table}

\begin{table}[H]
\caption{Prompt categories for generating the tuning section of OpenSQA. Each category guides GPT-4o to focus on a specific reasoning skill.}
\label{tab:tune_qatypes}
\begin{footnotesize}
\resizebox{\linewidth}{!}{
\begin{tabular}{|p{0.3\linewidth}|p{0.68\linewidth}|}
\hline
\cellcolor[HTML]{CBCEFB}\textbf{Reasoning Category} & \cellcolor[HTML]{CBCEFB}\textbf{Prompt Instruction} \\ \hline
\textbf{Scientific Reasoning} & Generate questions relating to physics or biomechanics explaining how IMU sensor dynamics align with activity type. \\ \hline
\textbf{Narrative Interpretation} & Describe activity based on temporal signal patterns, motion consistency, and event flow. \\ \hline
\textbf{Reliability Assessment} & Assess IMU data quality, consistency, and presence of noise or anomalies that might distort interpretation. \\ \hline
\rowcolor[HTML]{E8E8E8} \textbf{Prompt Input} & Activity label, IMU sensor data, sensor location, derived features \\
\rowcolor[HTML]{F5F5F5} \textbf{Output} & Five contextual question–answer pairs per reasoning type \\
\hline
\end{tabular}}
\end{footnotesize}
\end{table}

\begin{table}[H]
\caption{Prompt design for zero-shot activity classification from IMU signals.}
\label{tab:classification_prompt}
\begin{footnotesize}
\resizebox{\linewidth}{!}{
\begin{tabular}{|ll|}
\hline
\multicolumn{2}{|c|}{\cellcolor[HTML]{CBCEFB}\textbf{Instruction to LLM}} \\ \hline
\multicolumn{2}{|p{\columnwidth}|}{
Given IMU data, identify the activity class from the following options: \texttt{standing}, \texttt{sitting}, \texttt{walking}, \texttt{run}, \texttt{bike}, or \texttt{Unclear}. Return: \texttt{"The identified class is: <class\_name>"}.} \\ \hline
\cellcolor[HTML]{CBCEFB}\textbf{Input Modality} & \cellcolor[HTML]{CBCEFB}\textbf{Output Format} \\
\hline
\begin{tabular}[c]{@{}p{0.48\linewidth}@{}}
\textbf{LLaSA:} 194-token sensor embedding \\
\textbf{Other LLMs:} Raw sensor sequences (~2175 tokens)
\end{tabular}
&
\texttt{"The identified class is: <class\_name>"} \\ \hline
\end{tabular}}
\end{footnotesize}
\end{table}

\begin{table}[H]
\caption{Prompt for evaluating QA answer quality. GPT-4o scores LLM outputs using structured rubric.}
\label{tab:qa_eval_prompt}
\begin{footnotesize}
\resizebox{\linewidth}{!}{
\begin{tabular}{|ll|}
\hline
\multicolumn{2}{|c|}{\cellcolor[HTML]{CBCEFB}\textbf{Instruction to GPT-4o}} \\ \hline
\multicolumn{2}{|p{\columnwidth}|}{
Use the standard answer, activity label, sensor location, and predicted answer to evaluate correctness, completeness, consistency, and helpfulness. Provide a numeric score and brief assessment.} \\ \hline
\cellcolor[HTML]{CBCEFB}\textbf{Prompt Input} & \cellcolor[HTML]{CBCEFB}\textbf{Output Format} \\
\hline
\begin{tabular}[c]{@{}p{0.48\linewidth}@{}}
1. Standard answer \\
2. Activity label \\
3. Sensor location \\
4. Predicted answer
\end{tabular}
&
\begin{tabular}[c]{@{}p{0.48\linewidth}@{}}
\texttt{Quality score: <0–100>} \\
\texttt{Assessment: <1–2 sentence summary>}
\end{tabular} \\ \hline
\end{tabular}}
\end{footnotesize}
\end{table}

\section*{Appendix B: On the Non-Determinism of LLM Inference}

Our classification results are obtained with \textbf{LLaSA-7B} using the \textbf{LIMUBERT} adapter trained on four public IMU datasets (details in subsection \ref{limubert-training}) on 131,407 QAs derived from the four datasets, which can be repeated through our code and data available at \href{https://github.com/BASHLab/LLaSA}{https://github.com/BASHLab/LLaSA}. Although we run inference with apparently deterministic settings (temperature = 0, fixed prompt, fixed random seed, single-GPU inference), recent systematic measurements show that outputs can still vary across repeated runs. Atil et al. \cite{atil2024non} report accuracy variation up to 15\% across runs and gaps up to 70\% between best and worst runs for certain tasks under "deterministic" settings.

Such nondeterminism arises from subtle floating-point differences in GPU kernels (non-associativity, rounding), software backend changes (e.g., cuDNN/CUDA version differences), and runtime effects that perturb logits. While these effects are often very small numerically, they can flip the predicted class when logits are close, causing measurable changes in classification accuracy.

\textbf{Implication for Future Work.} For reproducible and fair comparisons, we encourage the community to adopt \emph{variance-aware evaluation}: run inference multiple times (e.g., $N=10$ runs), report mean or median accuracy, and include statistical dispersion metrics such as variance, best/worst range, and agreement scores. This would make reported results more robust and facilitate meaningful model comparisons.

Our own results are based on a single run per configuration, so we expect some variation if repeated. Incorporating multi-run statistics is an important direction for future work and would help characterize the stability of similar LLMs more systematically.




\end{document}